\documentclass{article}
\pdfoutput=1
\usepackage[nonatbib,preprint]{neurips_2021}

\usepackage[utf8]{inputenc} %
\usepackage[T1]{fontenc}    %
\usepackage{hyperref}       %
\usepackage{url}            %
\usepackage{booktabs}       %
\usepackage{amsfonts}       %
\usepackage{amssymb}
\usepackage{nicefrac}       %
\usepackage{microtype}      %
\usepackage[dvipsnames]{xcolor}
\usepackage{dirtytalk}
\usepackage{graphicx}
\usepackage{amsmath}
\usepackage{amssymb}
\usepackage{mathtools}
\usepackage{pifont}
\usepackage{placeins}
\usepackage{hyperref}
\usepackage{xspace}
\usepackage{wrapfig}
\usepackage{floatrow}
\usepackage{enumitem}
\usepackage{caption}
\usepackage{subcaption}
\newfloatcommand{capbtabbox}{table}[][\FBwidth]

\newcommand{\cmark}{\ding{51}}%
\newcommand{\xmark}{\ding{55}}%

\newcommand{\Model}{\emph{Frozen}\xspace}

\title{Multimodal Few-Shot Learning with \\ Frozen Language Models}

\author{%
Maria Tsimpoukelli$^{\ast}$\\
DeepMind\\
\texttt{mrts@deepmind.com}\\
\And
Jacob Menick$^{\ast}$\\
DeepMind\\
University College London\\
\texttt{jmenick@deepmind.com}\\
\And
Serkan Cabi$^{\ast}$\\
DeepMind\\
\texttt{cabi@deepmind.com}\\
\And
S. M. Ali Eslami\\
DeepMind\\
\texttt{aeslami@deepmind.com}\\
\And
Oriol Vinyals\\
DeepMind\\
\texttt{vinyals@deepmind.com}\\
\And
Felix Hill\\
DeepMind\\
\texttt{felixhill@deepmind.com}\\

}

\begin{document}

\maketitle

\begin{abstract}

When trained at sufficient scale, auto-regressive language models exhibit the notable ability to learn a new language task after being prompted with just a few examples. Here, we present a simple, yet effective, approach for transferring this few-shot learning ability to a multimodal setting (vision and language). Using aligned image and caption data, we train a vision encoder to represent each image as a sequence of continuous embeddings, such that a pre-trained, frozen language model prompted with this prefix generates the appropriate caption. The resulting system is a multimodal few-shot learner, with the surprising ability to learn a variety of new tasks when conditioned on examples, represented as a sequence of multiple interleaved image and text embeddings. We demonstrate that it can rapidly learn words for new objects and novel visual categories, do visual question-answering with only a handful of examples, and make use of outside knowledge, by measuring a single model on a variety of established and new benchmarks.

\end{abstract}

\section{Introduction}

Auto-regressive transformers have been shown to be very impressive models of natural language \cite{vaswani2017attention}. Large-scale language transformers exhibit several surprising abilities beyond that of standard text generation \cite{brown2020language,raffel2019exploring}. Perhaps most notably, they are \emph{few-shot learners}; they can learn to perform a new task from a few examples without any further gradient updates. Equipped with this ability, these models have been shown to rapidly adapt to new tasks and styles of generation via prompting (e.g.\ switching from formal to informal language)~\cite{brown2020language}, to quickly retrieve relevant encyclopedic or general knowledge when primed with a relevant context (e.g.\ answering questions such as `When did the French Revolution begin?') \cite{roberts2020much, meena, lama} and to use new words in appropriate ways straight after being taught what those words mean (sometimes referred to as `fast binding')~\cite{heibeck1987word,brown2020language}.

\begin{figure}[ht]
\centering
\includegraphics[width=\linewidth]{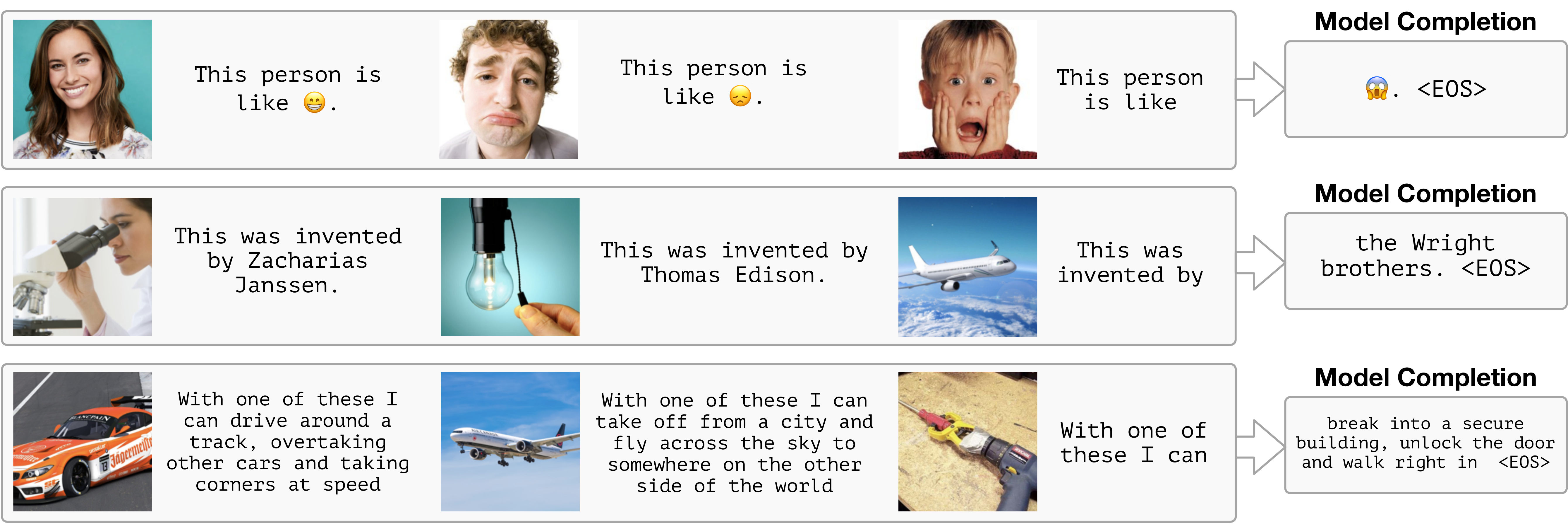}
\caption{Curated samples with about five seeds required to get past well-known language model failure modes of either repeating text for the prompt or emitting text that does not pertain to the image. These samples demonstrate the ability to generate open-ended outputs that adapt to both images and text, and to make use of facts that it has learned during language-only pre-training.}
\vspace*{-0.1cm}
\label{fig:headline}
\end{figure}

\begin{wrapfigure}[14]{r}{0.4\linewidth}
\centering
\vspace*{-0.65cm}
\includegraphics[width=\linewidth]{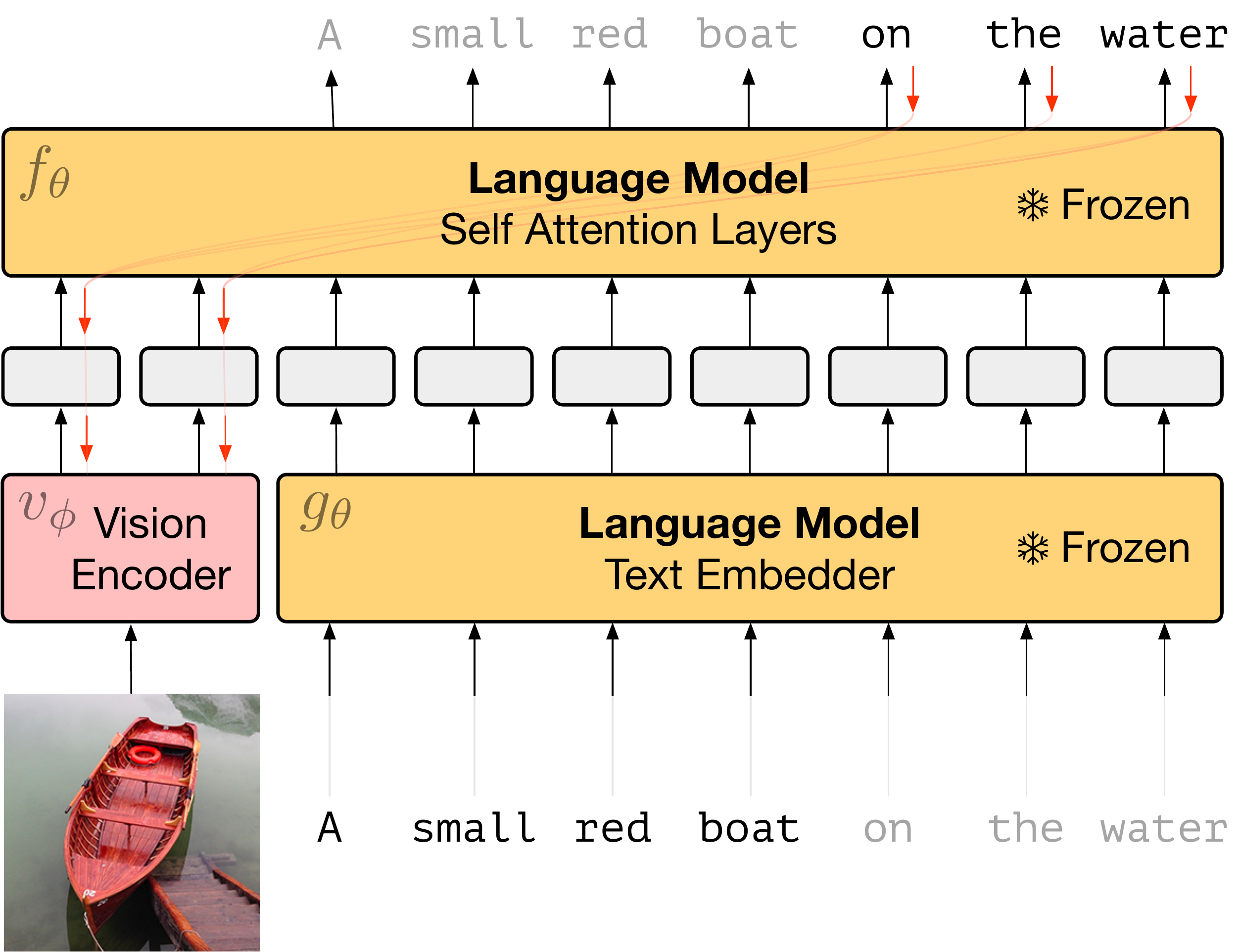}
\caption{Gradients through a frozen language model's self attention layers are used to train the vision encoder.}
\vspace*{-0.2cm}
\label{fig:method-training}
\end{wrapfigure}

Despite these impressive capabilities, such large scale language models are `blind' to modalities other than text, preventing us from communicating visual tasks, questions or concepts to them. Indeed, philosophers and linguists have questioned whether an un-grounded language model can ever achieve true understanding of the language it processes~\cite{chalmers,bender2020climbing}. Here, we present \Model, a method for giving a pre-trained language model access to visual information in a way that extends its few-shot learning capabilities to a multimodal setting, without changing its weights. \Model consists of a neural network trained to encode images into the word embedding space of a large pre-trained language model such that the language model generates captions for those images. The weights of the language model are kept frozen, but gradients are back-propagated \textit{through} it to train the image encoder from scratch (\autoref{fig:method-training}). Although \Model is trained on single image-text pairs, once trained it can respond effectively to ordered sets of multiple images and words. This allows users to e.g.\ `prompt' it with several examples of new multimodal tasks before evaluating its performance, or to `teach' it the name of a new visual category before immediately asking about that category. %

By exploiting its pre-trained language model, \Model exhibits strong zero-shot performance on multimdodal tasks that it was not trained on, such as visual question answering (VQA). More surprisingly, it gets better at these tasks after seeing a handful of examples ``in-context'' as in ~\cite{brown2020language}, and also performs above chance on tests of fast category learning such as miniImageNet~\cite{vinyals2016matching}. In each case, comparisons with `blind' baselines show that the model is adapting not only to the language distribution of these new tasks, but also to the relationship between language and images. \Model is therefore a \emph{multimodal few-shot learner}, bringing the aforementioned language-only capabilities of rapid task adaptation, encyclopedic knowledge and fast concept binding to a multimodal setting. 

Our goal in developing \Model was not to maximise performance on any specific task, and in many cases it is far from state-of-the-art. Nonetheless, it performs well above trivial baselines across a wide range of tasks without ever seeing more than a handful of the training examples provided by these benchmarks. Moreover, as illustrated in~\autoref{fig:headline}, \Model is a system for genuinely open-ended and unconstrained linguistic interpretation of images that often produces compelling output.

\begin{figure}[b!]
\vspace*{-0.7cm}
\centering
\includegraphics[width=\linewidth]{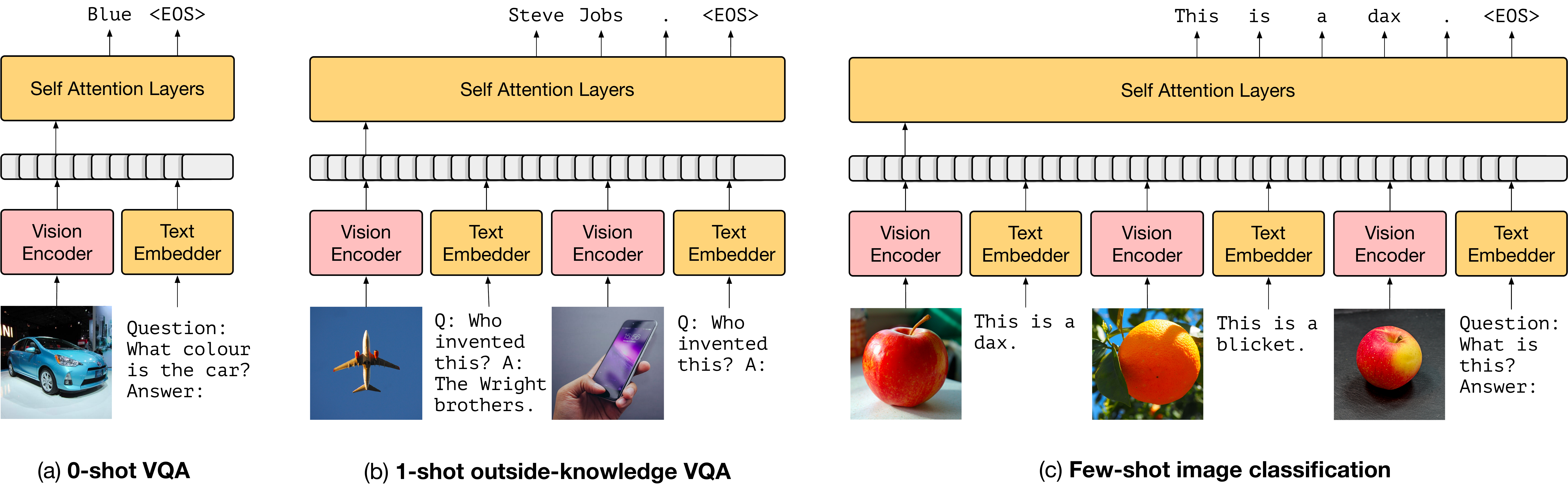}
\vspace*{-0.6cm}
\caption{Inference-Time interface for \Model. The figure demonstrates how we can support (a) visual question answering, (b) outside-knowledge question answering and (c) few-shot image classification via in-context learning.}
\label{fig:method-testing}
\end{figure}

To summarise, our contributions are as follows: 1.\ We present \Model, a modular, scalable and efficient approach to training vision front-ends for large language models. The resulting combined model retains all of the capabilities of large language models, but can also process text and image inputs in any arbitrary sequence. 2.\ We show that such models transfer their capacity for rapid task adaptation, encyclopedic knowledge and fast concept binding from a language-only to a multimodal setting, and verify that prompting them with both visual and language information can be strictly more effective than doing so with language information alone. 3.\ We quantify these capabilities on a range of existing and new benchmarks, paving the way for future analysis of these capabilities.

\section{Related Work}
The \Model method is inspired by lots of recent work.~\cite{lu2021pretrained} show that the knowledge encoded in transformer language models can be a valuable prior for tasks involving reasoning and memory across discrete sequences, and even classification of images presented as sequences of spatial regions. In that approach, a small subset of the pre-trained language model weights are fine-tuned to the various final applications. In contrast, applying \Model to different tasks does not involve any weight updates to the transformer whatsoever; the system adapts to and improves at multimodal (vision and language) tasks as activations propagate through the model. The two studies thus reveal different ways in which knowledge acquired from text can transfer to non-linguistic settings.

The effectiveness of \emph{prefix tuning} \cite{li2021prefix} or \emph{prompt tuning} \cite{lester2021power} was another important motivation for \Model. Prefix tuning is a method for prompting a language model to produce output of a particular style using gradient descent to learn a task-specific bias term which functions like the continuous embedding of a text prompt. Using prefix tuning, language models can be adapted to different natural language generation tasks like summarization. \Model could also be considered a type of \emph{image-conditional prefix tuning}, in which this continuous prompt is not a bias but an \emph{image-conditional} activation produced by an external neural network.

A large body of work has applied either text-specific or multimodal representation-learning approaches like BERT~\cite{devlin2018bert} to visual question answering (VQA) and captioning (see e.g.~\cite{lu2019vilbert,su2019vl} and many more). In these approaches, models are first trained with aligned data on task-agnostic cross-modal objectives and then fine-tuned to specific tasks. This approach can yield state-of-the-art performance on a range of classification tasks. Unlike \Model, the resulting systems are highly specialized to one task, and cannot learn new concepts or adapt to new tasks in a few shots. 

By contrast, \cite{cho2021unifying} propose text generation as an objective for task-general multimodal models, yielding a system that, like \Model, produces unconstrained language output. Unlike \Model, they do not use a pre-trained model trained on text only, and do not consider zero or few-shot learning, instead updating all weights of the system with training data for each task they consider -- thus, again, specializing the models to one task at a time. Similarly, \cite{ziegler2019encoder} and \cite{chen2021visualgpt} show that a large pre-trained language model as decoder can improve a captioning performance when training data is limited. Unlike \Model, they use pre-trained frozen visual encoders or object extractors and fine-tune the pre-trained weights in the text decoder on the captioning data. Similarly, they do not consider zero or few-shot adaptation across different multimodal tasks. Past work has also explored alternative approaches for post-hoc combination of models for different modalities using latent variables \cite{tian2019latent}.

Multimodal pre-training has recently been shown to enable strong zero-shot generalization in the discriminative setting using large-scale contrastive learning \cite{radford2021learning, jia2021scaling}. Also in a discriminative setting, \cite{zhai2021scaling} has observed signs of emergent few-shot-learning from large-scale training. In contrast, our work enables strong generalization to new multimodal tasks both zero-shot or few-shot with completely open-ended generative text output.

\section{The \Model Method}
\Model is a method for grounding a large language model without changing its weights, closely related to \emph{prefix tuning} \cite{li2021prefix,lester2021power}. Prefix tuning trains a task-specific continuous bias term to function like the embedding of a constant, static text prompt used for all test-time examples. \Model extends this approach by making this prefix \emph{dynamic}, in that it is not a constant bias but an input-conditional \emph{activation} emitted by a neural network.

\subsection{Architecture}
\paragraph{Pre-trained Autoregressive Language Models} 
Our method starts from a pre-trained deep auto-regressive language model, based on the Transformer architecture \cite{vaswani2017attention,Radford2019LanguageMA}, which parametrizes a probability distribution over text $\mathbf{y}$. Text is decomposed into a sequence of discrete tokens $\mathbf{y} = y_1, y_2, ..., y_L$ by the SentencePiece tokenizer \cite{kudo2018sentencepiece}. We use a vocabulary of size 32,000. The language model makes use of an embedding function $g_\theta$ which independently transforms each token into a continuous embedding $t_l \coloneqq g_\theta(y_l)$, as well as a transformer neural network $f_\theta$ whose output is a vector of logits parameterizing a categorical distribution over the vocabulary. The distribution $p_{\theta}(\mathbf{y})$ is represented as follows:
\begin{align*} 
\log p_{\theta}(\mathbf{y}) &= \sum \limits_{l} \log p_{\theta}(y_l|y_1, y_2, ..., y_{l - 1})
                 = \sum \limits_{l} f_{\theta}(t_1, t_2, ..., t_{l - 1})_{y_l}
\end{align*}
The model we start from is pre-trained, i.e. $\theta$ has been optimised via the standard maximum-likelihood objective on a large dataset of text from the internet. We use a 7 billion parameter transformer trained on the public dataset C4 \cite{raffel2019exploring} -- previous work has shown that the multi-billion parameter scale is sufficient to exhibit the key capacities we are interested in studying \cite{Radford2019LanguageMA, roberts2020much}.

\paragraph{Vision Encoder} Our vision encoder is based on NF-ResNet-50 \cite{brock2021high}. We define $v_{\phi}$ as a function that takes a raw image and emits a continuous sequence to be consumed by the transformer. We use the final output vector of the NF-Resnet \emph{after} the global pooling layer.

\paragraph{Visual Prefix}
One important requirement is to represent images in a form that the transformer \emph{already} understands: a sequence of continuous embeddings, each having the same dimensionality $D$ as a token embedding $t_l$. We therefore form the visual prefix by linearly mapping the vision encoder's output to $D * n$ channels, and then reshaping the result as a sequence of $n$ embeddings, each with dimensionality $D$.  We call this sequence a \emph{visual prefix} since it plays the same functional role in the transformer architecture as (part of) an embedding sequence of prefix tokens. We experimented using different number of tokens, specifically 1, 2 and 4 and found that 2 performs best, though certainly this would be sensitive to other architectural details. See Appendix for more details on the architecture.

\subsection{Training}
During training, we update only the parameters $\phi$ of the vision encoder using paired image-caption data from the Conceptual Captions dataset \cite{sharma2018conceptual}. Our experiments show that fine-tuning $\theta$ hurts generalization, as much less paired image-caption data is available than the amount of text-only data used to pre-train $\theta$. Training only the parameters $\phi$ makes our system \emph{modular} -- it can use an existing language model off the shelf -- and also quite simple: we only train a visual encoder and rely on the capabilities of an existing language model.

Following standard captioning systems \cite{li2019visual, hossain2019comprehensive}, we treat captioning as conditional generation of caption text $\mathbf{y}$ given an image $\mathbf{x}$. We represent $\mathbf{x}$ as $v_{\phi}(\mathbf{x}) = i_1, i_2, ..., i_n$ and train $\phi$ to maximise the likelihood:
\begin{align*} 
\log p_{\theta, \phi}(\mathbf{y} | x) &= \sum \limits_{l} \log p_{\theta, \phi}(y_l|\mathbf{x}, y_1, y_2, ..., y_{l - 1}) \\
                  &= \sum \limits_{l} f_{\theta}(i_1, i_2, ..., i_n, t_1, t_2, ..., t_{l - 1})_{y_l}
\end{align*}
Whilst the parameters $\theta$ are frozen, each element $i_k$ of the visual prefix receives gradients $\sum \limits_{l} \nabla_{i_k} f_{\theta}(i_1, i_2, ..., i_n, t_1, t_2, ..., t_{l - 1})_{y_l}$, enabling the parameters of the visual encoder to be optimised with standard backpropagation and SGD (\autoref{fig:method-training}).

As the notation $f_{\theta}(i_1, i_2, ..., i_n, t_1, t_2, ..., t_{l - 1})$ suggests, we present the visual prefix during training as if it were a sequence of embeddings occurring earlier in time than the caption (token embeddings) $t_1, t_2, ...$. We use relative positional encoding \cite{relative}, which enables the transformer to generalize to prompt sequences where an image is not always in the first absolute positions, and where more than one image may be present. We leave improvements of this simple scheme for future work.

\subsection{Interface at Inference Time}

At inference time, a vanilla language model, conditioned upon an arbitrary text prompt or `prefix' $y_1, y_2, ..., y_p$, generates text sequences $y_{p+1}, y_{p+2}, ...$ autoregressively. In \Model it is straightforward to include images in a prompt by placing an image's embedding $i_1, i_2$ next to a text embedding subsequence $t_1, t_2, ..., t_p$. Because the transformer $f_{\theta}$ is modality-agnostic, we can interleave a sub-sequence of text token embeddings with a sub-sequence of image embeddings in any arbitrary order. In \autoref{fig:method-testing}, we show how this can support zero-shot visual question-answering (\autoref{fig:method-testing}a), few-shot visual question-answering (\autoref{fig:method-testing}b), and few-shot image classification (\autoref{fig:method-testing}c).

To evaluate these tasks, the model \emph{decodes} output sequences greedily and these outputs are compared against the ground truth answers of the task following the normalization technique used in \cite{VQAGithub}. We do not use short-lists of pre-canned answers to stress test the open-ended capabilities of \Model, even though in some tasks this may hurt its performance.

\begin{figure}[t!]
\includegraphics[width=\linewidth]{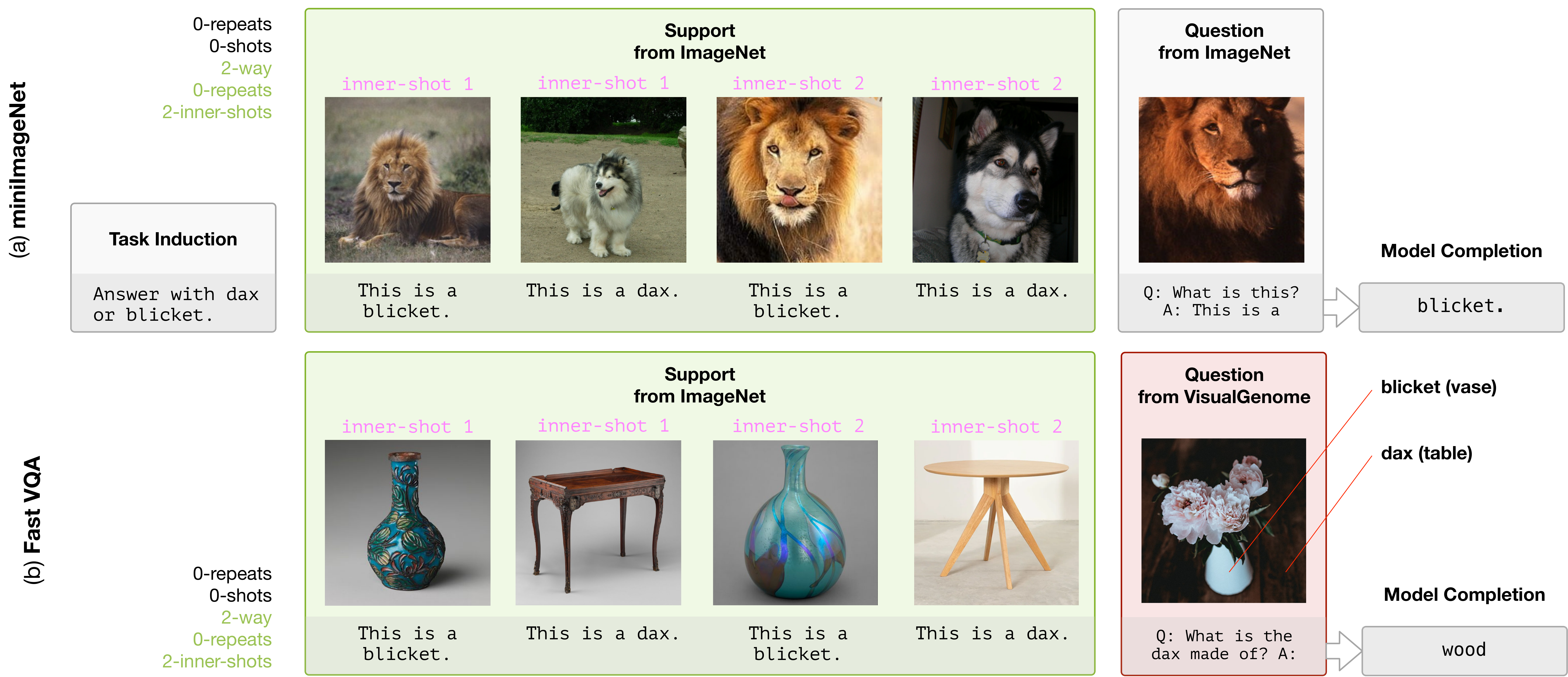}
\caption{Examples of (a) the Open-Ended miniImageNet evaluation (b) the Fast VQA evaluation.}
\label{fig:vocab_mi_fvqa}
\vspace*{-0.1cm}
\end{figure}

\subsection{Few-Shot Learning Definitions}

The ability of \Model to be conditioned on a sequence of interleaved images and text allows it not only to be able to perform at different multimodal tasks, but also gives rise to different ways of `inducing' the task to the model in order to improve its performance. We briefly define the terminology used in our settings, common amongst all the different tasks. See \autoref{fig:vocab_complete} in the appendix for a visual illustration of these concepts.

\begin{itemize}[]
    \item \textbf{Task induction} Explanatory text that precedes the sequence of images and text. It is intended to describe the task to the model in natural language, for example `Please answer the question.'
    \item \textbf{Number of shots} The number of distinct full examples of the task presented to the model prior to the evaluated example. For example, in Visual Question-Answering, a shot is an image along with the question and the answer.
\end{itemize}

For tasks involving fast concept binding (e.g., few-shot image classification), we define further specific terminology. See also \autoref{fig:vocab_mi_fvqa}a and \autoref{fig:vocab_extended_complete} in the appendix.
\begin{itemize}[]
    \item \textbf{Number of ways} The number of object classes in the task (e.g.\ dog vs cat).
    \item \textbf{Number of inner-shots} The number of distinct exemplars from each category that are presented to the model (i.e.\ number of images of different dogs). In previous work with MiniImagenet, these were known as \emph{shots}, but we modify the term here to distinguish from the more general usage of the term described above. 
    \item \textbf{Number of repeats} The number of times each inner-shot is repeated in the context presented to the model. We use this setting as an ablation to explore how the model integrates visual information about a category.
\end{itemize}

\section{Experiments: A Multi-Modal Few-Shot Learner}
\label{sec:experiments}
Our experiments are designed to quantify three capacities that should be characteristic of a Multi-Modal Few-Shot Learner: \emph{rapid adaptation} to new tasks, fast access to~\emph{general knowledge} and \emph{fast binding} of visual and linguistic elements. We train \Model on Conceptual Captions, a public dataset that consists of around three million image-caption pairs \cite{sharma2018conceptual}. We do early stopping on the validation set perplexity which usually reaches an optimum just after a single epoch with batch size 128. All experiments used the Adam optimizer with $\beta_1 = 0.9$ and $\beta_2 = 0.95$ and a constant learning rate of $3e\text{-}4$ unless otherwise noted. We operate on 224$\times$224 images at both train and test-time. Images which are not square are first padded with zeroes to square and then resized to 224$\times$224.

\subsection{Rapid Task Adaptation}
\label{sec:rapid_adaptation}

We first examine zero-shot and few-shot generalization from captioning to visual question-answering. This is a type of \emph{rapid adaptation} from captioning behaviour to question-answering behaviour with either simple prompting alone or few-shot learning, analogous to transfer from language modelling to open-domain question-answering \cite{roberts2020much} in the vision plus language domain. We evaluate on the VQAv2 \cite{goyal2017making} validation set.

\paragraph{Zero-shot transfer from captioning to VQA}

Captioning training can transfer moderately well to visual question-answering in the zero-shot setting with no training or in-context examples at all. The strength of the pre-trained language model is a double-edged sword. It powers the generalization abilities of \Model but also enables the model to perform surprisingly well without considering the visual input at all. To guard against this possibility we also train blind baselines, in which the image presented to the visual encoder is blacked out, but the convnet weights are still trained. This amounts to prefix tuning \cite{li2021prefix}. We outperform this blind baseline which also inherits the few-shot learning abilities of the language model.

In these experiments we also include two additional and important baselines: $\text{\Model}_{\,\text{finetuned}}$ in which the language model is instead finetuned starting from the pretrained weights and $\text{\Model}_{\,\text{scratch}}$, wherein the whole system is trained from scratch end-to-end. These baselines preferred a smaller learning rate of $1e\text{-}5$. Results in \autoref{tab:vqa_table} show that keeping the language model frozen generalizes substantially better to visual question-answering than finetuning. The model trained from scratch is not able to transfer at all from captioning to VQA; we interpret this to suggest that the tremendous generalization abilities of large language models are reliant upon large-scale training datasets in which the task of predicting the next token mimics the test setting (here question-answering) with non-negligible frequency.

\begin{table}
\begin{floatrow}
\hspace*{-0.25cm}
\capbtabbox{
\begin{tabular}{l||c|c|c|c}
\textbf{n-shot Acc.}     & \textbf{n=0} & \textbf{n=1} & \textbf{n=4} & $\tau$ \\ \toprule
\textbf{\Model} & 29.5 & 35.7 & 38.2 & \xmark \\
\textbf{\Model$_{\text{scratch}}$} & 0.0 & 0.0& 0.0 & \xmark \\
\textbf{\Model$_{\text{finetuned}}$} & 24.0& 28.2& 29.2 & \xmark \\
\textbf{\Model$_{\text{train-blind}}$} &26.2 &33.5 & 33.3 & \xmark \\
\midrule
\textbf{\Model$_{\text{VQA}}$} &48.4 & -- & -- & \cmark \\
\textbf{\Model$_{\text{VQA-blind}}$} & 39.1 &  -- &   -- & \cmark \\
\midrule
\textbf{Oscar \cite{oscar}}   &            73.8 &         -- &         -- & \cmark \\ \bottomrule
\end{tabular}
}{
\caption{Transfer from Conceptual Captions to VQAv2. The $\tau$ column indicates whether a model uses training data from the VQAv2 training set. The row denoted \Model$_{\text{train-blind}}$ is the blind baseline described in \autoref{sec:rapid_adaptation}. \Model$_{\text{VQA}}$ is a baseline which mixes in VQAv2 training data.}
\label{tab:vqa_table}
}
\capbtabbox{
\begin{tabular}{l||c|c|c|c}
\textbf{n-shot Acc.} & \textbf{n=0} & \textbf{n=1} & \textbf{n=4} & $\tau$ \\ \toprule
\textbf{\Model} & 5.9 & 9.7 & 12.6 & \xmark \\
\textbf{\Model$_{\text{400mLM}}$} & 4.0 & 5.9 & 6.6 & \xmark \\
\textbf{\Model$_{\text{finetuned}}$} & 4.2 & 4.1 & 4.6 & \xmark \\
\textbf{\Model$_{\text{train-blind}}$} & 3.3 & 7.2 & 0.0 & \xmark \\
 \midrule
\textbf{\Model$_{\text{VQA}}$} & 19.6 & -- & -- & \xmark \\
\textbf{\Model$_{\text{VQA-blind}}$} & 12.5 & -- & -- & \xmark \\
\midrule
\textbf{MAVEx \cite{wu2021multimodal}} & 39.4 & -- &-- & \cmark \\
\bottomrule
\end{tabular}
}{
\caption{Transfer from Conceptual Captions to OKVQA. The $\tau$ column indicates if a model uses training data from the OKVQA training set. \Model does not train on VQAv2 except in the baseline row, and it never trains on OKVQA.}
\label{tab:okvqa_table}
}
\end{floatrow}
\end{table}

\paragraph{Improving performance with few-shot learning}

This zero-shot transfer to visual question-answering via prompting improves by presenting examples to the model in-context. We repeat the previous experiments with up to four examples of image-question-answer triples shown to the model as conditioning information in the continuous prompt sequence (using the interface in \autoref{fig:method-testing}).  We present these few-shot results compared to mixing in data from the VQAv2 training set -- for SGD training -- in \autoref{tab:vqa_table}. Of course, few-shot learning on four examples is outperformed by SGD on tens of thousands of examples, but few-shot performance clearly improves with more examples and goes a decent way toward closing the gap from zero-shot performance (29.5\%) to full SGD training performance (48.4\%). With just four examples the gap is closed almost halfway at 38.2\%.

There are two important takeaways from the results presented in this section. First, they show that training a visual encoder through a pretrained and frozen language model results in a system capable of strong out-of-distribution (zero-shot) generalization. Second, they confirm that the ability to rapidly adapt to new tasks given appropriate prompts is inherited from the pretrained language model and transfers directly to multimodal tasks.

\subsection{Encyclopedic Knowledge}\label{sec:encyc_knowledge}

Here we study the extent to which \Model can leverage the encyclopedic knowledge in the language model towards visual tasks. The Conceptual Captions dataset is \emph{hypernymed} meaning that e.g.\ proper names are replaced with a general word like {\it person}. This enables us to rigorously study the transfer of factual knowledge because all knowledge of named entities comes from language model pretraining.

Consequently, when we show the model an image of an airplane and ask ``who invented this?'' (\autoref{fig:headline}), the visual encoder has determined that the image contains an airplane, and the language model has used this to retrieve the factual knowledge that airplanes were invented by the Wright brothers, a fact which is referenced in the C4 training set through (text-only) articles about airplanes. This is a fascinating chain of deduction. A detailed analysis of this behaviour with more examples is included in the Appendix (e.g.\ \autoref{fig:knowledge_wine}, \autoref{fig:knowledge_emoji}, \autoref{fig:wine_really}).

We bolster this finding quantitatively by evaluating performance on OKVQA \cite{marino2019ok}, a visual question-answering dataset designed to require outside knowledge in order to answer correctly. The pretrained language model's command of factual knowledge is of course dependent upon its scale, so we examine the performance of \Model using pretrained language models of varying sizes: the base model with 7 billion parameters, and a much smaller 400 million parameter language model pretrained on the same dataset. \autoref{tab:okvqa_table} shows the results: task performance scales with model size. Again finetuning performs worse than leaving the model frozen in terms of generalization performance. We stress that \Model is never trained on OKVQA.

\subsection{Fast Concept Binding}
In the multi-modal setting, fast-binding refers to a model's ability to associate a word with a visual category in a few shots and immediately use that word in an appropriate way.

\paragraph{Open-Ended miniImageNet and Real-Name miniImageNet}

To quantify the fast-binding capacity of of \Model, we evaluate it on the minImageNet meta-learning task \cite{vinyals2016matching}. Note that there are important differences with how we attempt miniImageNet and how it is approached in previous work. First, unlike standard meta-learning, we do not train \Model on the (meta) task. Second, we evaluate \Model in an open-ended fashion, where it must successfully generate a correct category name (and then the EOS token) in order to be credited with a correct answer. Finally, although we use the same image classes as the miniImageNet test set, they are at higher resolution (224$\times$224) and with class labels replaced with nonsense words (`dax', `blicket' etc). This allows the system to express its answers with word-like tokens. We refer to this task as \emph{Open-Ended miniImageNet}, and it mimics closely the standard miniImagenet setting used elsewhere. To assess how much difficulty is added by binding visual categories to nonsense words versus simply adapting to an image recognition task \emph{per se}, we also consider a version --  Real-Name miniImagenet -- in which visual categories in both the support set and the answer retain their original names. See \autoref{fig:vocab_mi_fvqa}a for an illustration. 

On both versions of this evaluation, we experiment by exposing the model to different numbers of inner-shots, repeats and task induction. On two-way Open-Ended miniImagenet, we observe that when \Model is presented with a sequence of images and descriptions of new names for them, it is able to learn new names for the objects presented and then use these new names immediately with substantially above chance accuracy. Importantly, the ability of the model to use these new words improves with with more examples of the corresponding category. Notably, this upward trend is more pronounced when this supporting information involves different exemplars from the visual category (inner-shots) rather than repetitions of a single exemplar (repeats). The fast-binding capacities of the model can thus be improved with richer and more varied visual support or prompting.

On two-way Real-Name miniImagenet, we observe a similar trend but with higher absolute performance. This underlines the difficulty in Open-Ended miniImagenet introduced by having to assign novel words to categories that may otherwise be already known to the model, and because the real names may carry visual information leveraged from the captioning data the model was trained on. 

In \autoref{tab:mi5}, we show that the observed effects on Open-Ended miniImagenet do not transfer to the 5-way setting, where \Model is not significantly above chance. This shows that learning to bind five new names to five visual categories in a single forward pass is beyond the current capabilities of \Model. As before, however, we do observe an upward trend in the model's capacity to return the actual name for a visual category among the five possibilities as the number of inner-shots or repeats increases. Further work is required and we look forward to progress in this more challenging setting.

\begin{table}[h]
\centering
\begin{tabular}{l||l|lll|lll}
\textbf{Task Induction}                         & \xmark    & \cmark    & \cmark             & \cmark             & \cmark    & \cmark             & \cmark             \\
\textbf{Inner Shots}                            & 1    & 1    & 3             & 5             & 1    & 1             & 1             \\
\textbf{Repeats}                          & 0    & 0    & 0             & 0             & 1    & 3             & 5             \\ \midrule
\textbf{\Model}                 & 29.0 & 53.4 & 57.9 & 58.9 & 51.1 & 57.7 & 58.5 \\
\textbf{\Model (Real-Name)}     & 1.7  & 33.7 & 66   & 66   & 63   & 65   & 63.7          \\ \midrule
\textbf{\Model$_{\text{test-blind}}$}    & --   & 48.5 & 46.7          & 45.3          & --   & --            & --            \\
\textbf{\Model$_{\text{test-blind}}$ (Real-Name)} & --   & 1.0  & 12.6          & 33.0          & --   & --            & --            \\
\textbf{ANIL Baseline \cite{raghu2019rapid}}                                   & --   & 73.9 & 81.7          & 84.2          & --   & --            & -- \\ \bottomrule          
\end{tabular}
\vspace{0.cm}
\caption{Performance of \Model and baselines on Open-Ended miniImageNet 2-Way Tasks. Randomly picking between the two class labels (then emitting the EOS token) would yield 50\% accuracy. As the model has to generate the answer, and is not counted correct if it paraphrases, this is not the best blind baseline, which is why we include open-ended blind baselines that also generate.}
\label{tab:mi2}
\end{table}

\begin{table}[h]
\centering
\begin{tabular}{l||l|lll|lll}
\textbf{Task Induction}                                  & \xmark    & \cmark    & \cmark             & \cmark             & \cmark             & \cmark             & \cmark    \\
\textbf{Inner Shots}                                     & 1    & 1    & 3             & 5             & 1             & 1             & 1    \\
\textbf{Repeats}                                   & 0    & 0    & 0             & 0             & 1             & 3             & 5    \\ \midrule
\textbf{\Model}                          & 18.0 & 20.2 & 22.3 & 21.3 & 21.4 & 21.6 & 20.9 \\
\textbf{\Model (Real-Name)}              & 0.9  & 14.5 & 34.7 & 33.8 & 33.8 & 33.3 & 32.8 \\ \midrule
\textbf{\Model$_{\text{test-blind}}$}             & --   & 18.6 & 19.9          & 19.8          & --            & --            & --   \\
\textbf{\Model$_{\text{test-blind}}$ (Real-Name)} & --   & 4.6  & 22.6          & 20.8          & --            & --            & --   \\
\textbf{ANIL Baseline \cite{raghu2019rapid}}                                                     & --   & 45.5 & 57.7          & 62.6          & --            & --            & --  \\ 
\bottomrule
\end{tabular}
\caption{Performance of \Model and baselines on Open-Ended miniImageNet 5-Way Tasks. Randomly picking between the five class labels (then emitting the EOS token) would yield 20\% accuracy.}
\label{tab:mi5}
\vspace*{-0.1cm}
\end{table}

\paragraph{Fast-VQA and Real-Fast-VQA}

As transformers are trained to model text, their attention weights learn to associate -- or `bind'-- pairs of words across sentences. The experiments with miniImageNet show that this capacity can transfer directly to binding visual categories to their names, enabling the system to generate the name on demand. This raises the question of whether \Model can integrate a newly-acquired visual category (and its names) more fully into the model's language system, so that it can, for instance, describe or answer questions about that  category. 

To test this capacity, we constructed a new task -- \emph{Fast-VQA} -- out of two well-known datasets, ImageNet \cite{russakovsky2015imagenet} and Visual Genome \cite{krishna2017visual}. For each question, the model is presented with nonsense words (`dax' and `blicket') and $n$ images of the referents of those words (e.g.\ of a `cat' or a `dog') taken from ImageNet. It is then asked a question containing at least one of those two words, about a further image (taken from Visual Genome) in which \emph{both} of the referents appear (see \autoref{fig:vocab_mi_fvqa}b). As with miniImagenet, the words `dax' and `blicket' (and how they refer) should be new to \Model, but the corresponding visual categories may be known from the Conceptual Captions training data, albeit by different names.

To quantify how much harder the introduction of new words for known categories makes this task, we also created a variant (\emph{Real-Fast-VQA}) in which the original category names (`cat' or `dog') are used instead of `dax' and `blicket'. Real-Fast-VQA is a special case of VQA involving questions from Visual Genome, in which a model is reminded what the important entities in the question look like prior to answering the question. \emph{Real-Fast-VQA} does not require the same ability to bind categories to new words, but it does measure how well a model can exploit task-relevant multimodal guidance when attempting a new task in an otherwise zero-shot manner.

Fast-VQA and Real-Fast-VQA are very challenging tasks because they are attempted without task-specific training, and because the underlying questions come from Visual Genome (VQAv2 images do not come with the necessary meta-data to construct the task). Visual Genome questions are particularly challenging because only a single answer exists for each question. When scoring models, for simplicity we credit only an exact match with the output generated by the model, modulo the same post-processing applied for VQAv2. Because of the inherent difficulty of the task, we use strong baselines to verify strength of observed effects. The Fast-VQA and Real-Fast-VQA evaluation sets will be provided with the camera ready version of this manuscript, as a resource to stimulate further research on multimodal fast-binding, together with training data (not used in this work).

\begin{table}[h]
\centering
\begin{tabular}{l||llllllll}
\multicolumn{1}{l}{}                 & \multicolumn{4}{c}{\bf Fast-VQA}     & \multicolumn{4}{c}{\bf Real-Fast-VQA}        \\
\multicolumn{1}{l||}{\bf Inner Shots}       & 0   & 1   & 3   & \multicolumn{1}{l|}{5}   & 0   & 1   & 3    & \multicolumn{1}{l}{5}    \\ \toprule
\multicolumn{1}{l||}{\bf \Model}       & 1.6 & 2.8 & 7.0 & \multicolumn{1}{l|}{7.9} & 3.7 & 7.8 & 10.1 & \multicolumn{1}{l}{10.5} \\
\multicolumn{1}{l||}{\bf \Model$_{\text{train-blind}}$} & 0.7 & 0.3 & 1.3 & \multicolumn{1}{l|}{0.4} & 1.9 & 2.3 & 3.7  & \multicolumn{1}{l}{3.7} \\
\bottomrule
\end{tabular}
\caption{Performance of \Model versus an equivalent blind model on Fast and Real-Fast VQA.}
\label{tab:dax-vqa-results}
\end{table}

As shown in \autoref{tab:dax-vqa-results}, the fact that the model improves with more shots in both Fast-VQA and Real-Fast-VQA confirms that \Model has some capacity to integrate novel words into its general capacity to process and generate natural language in a multimodal context. It is notable that a prefix-tuned model with no access to images improves moderately at Real-Fast-VQA as more concepts are presented, showing that additional linguistic cues (just being reminded of the words involved and the linguistic form of the task) goes some way to preparing for the upcoming question. As exemplified in \autoref{fig:vocab_mi_fvqa}, inspection of the model output confirms that in many cases it is indeed the multimodal (and not just linguistic) support that enables \Model to improve performance as the number of shots increases.

\section{Discussion}\label{conclusion}

\subsection{Limitations} We believe this work is an important proof-of-concept for a desired, much more powerful system capable of open-ended multimodal few-shot learning. \Model achieves the necessary capacities \emph{to some degree}, but a key limitation is that it achieves far from \emph{state-of-the-art} performance on the specific tasks that it learns in a few shots, compared to systems that use the full training set for those tasks. As such, the main contribution of this work should be seen as a starting point or baseline for this exciting area of research of multimodal few-shot learning.

Further improvement can make the impressive zero-shot and few-shot generalization we observed more robust as reflected by higher accuracy and fewer seeds required to demonstrate our most compelling samples. Finally, there are many technical questions that were not explored in this proof-of-concept study, such as whether performance could be improved with more elaborate architectures for mixing vision and language. We leave the exploration of these possibilities to future investigations. The Open-Ended miniImageNet, Real-Name miniImagenet, Fast-VQA and Real-Fast-VQA benchmarks that we will provide with the camera ready version of this manuscript should facilitate the evaluation and analysis of future systems of this type.

\subsection{Conclusion}We have presented a method for transforming large language models into multimodal few-shot learning systems by extending the soft-prompting philosophy of \emph{prefix tuning} \cite{li2021prefix} to ordered sets of images and text while preserving text prompting abilities of the language model. Our experiments confirm that the resulting system, \Model, is capable both of open-ended interpretation of images and genuinely multimodal few-shot learning even though the system is only trained to do captioning. One corollary of these results is that the knowledge required to quickly bind together or associate different words in language is also pertinent to rapidly binding language to visual elements across an ordered set of inputs. This finding extends the conclusion of \cite{lu2021pretrained} -- that knowledge in transformer language models can transfer to non-linguistic tasks -- to the specific case of knowledge about few-shot learning.

\paragraph*{Acknowledgements} We wish to thank Sebastian Borgeaud and Jack Rae for preparing the pretraining text dataset and pretraining a selection of transformer language models, as well as Trevor Cai for help with experiments and infrastructure. We also wish to thank Pauline Luc, Jeff Donahue, Malcolm Reynolds, Andy Brock, Karen Simonyan, Jean-Baptiste Alayrac, Antoine Miech, Charlie Nash, Aaron van den Oord, Marc Deisenroth, Aida Nematzadeh, Roman Ring, Francis Song, Eliza Rutherford, Kirsty Anderson, Esme Sutherland, Daan Wierstra, and Nando de Freitas for insightful discussions during the course of the project.

\bibliographystyle{plain}
\bibliography{references}

\clearpage
\FloatBarrier

\clearpage
\FloatBarrier

\appendix
\section{Appendix}
\subsection{Compute Usage}
The seven billion parameter language model we used as part of \Model used model parallelism with the strategy from \cite{shoeybi2020megatronlm} to partition one instance of the model over four accelerators. Each instance had a batch size of 8. To reach a batch size of 128 in this configuration, we additionally employed data parallelism with 16 synchronous replicas. The whole system was trained on a 4x8 TPUv3 \cite{jouppi} topology for about 12 hours, which is when validation set performance for Conceptual Captions led us to do early stopping.

\subsection{Frozen Architecture Details}

The pretrained transformer language model we used has a GPT-like architecture \cite{Radford2019LanguageMA}. It consists of a series of identical residual layers, each comprised of a self-attention operation followed by a positionwise MLP. The only deviation from the architecture described as GPT-2 is the use of relative position encodings \cite{relative}. Our seven billion parameter configuration used 32 layers, with each hidden layer having a channel dimensionality of 4096 hidden units. The attention operations use 32 heads each with key/value size dimensionality of 128, and the hidden layer of each MLP had 16384 hidden units. The 400 million parameter configuration used 12 layers, 12 heads, hidden dimensionality of 1536, and 6144 units in the MLP hidden layers.

\subsection{Few-Shot Learning Definitions}
\begin{figure}[ht!]
\includegraphics[width=\linewidth]{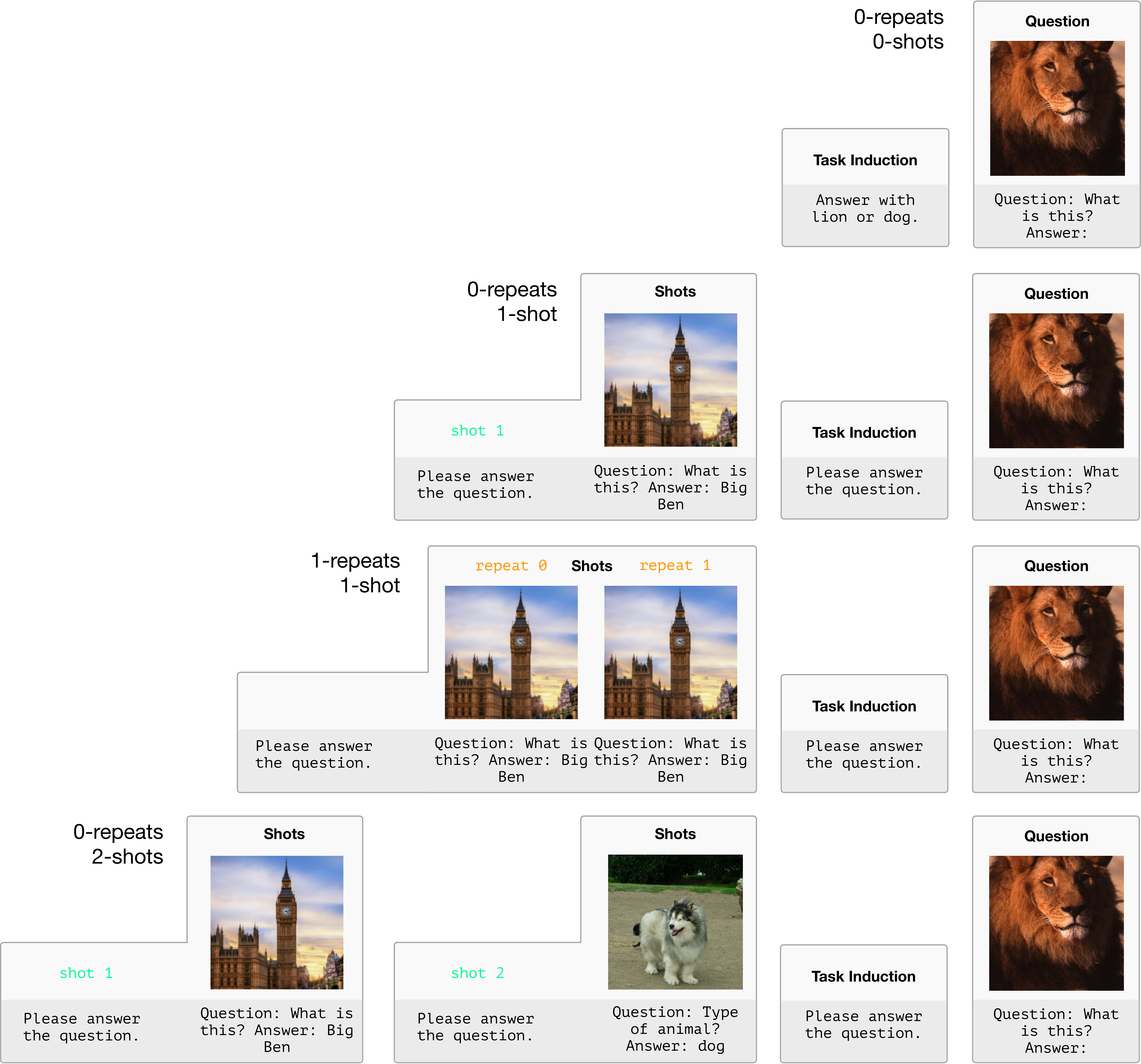}
\caption{Examples of few-shot learning vocabulary.}
\label{fig:vocab_complete}
\end{figure}

\begin{figure}[ht!]
\includegraphics[width=\linewidth]{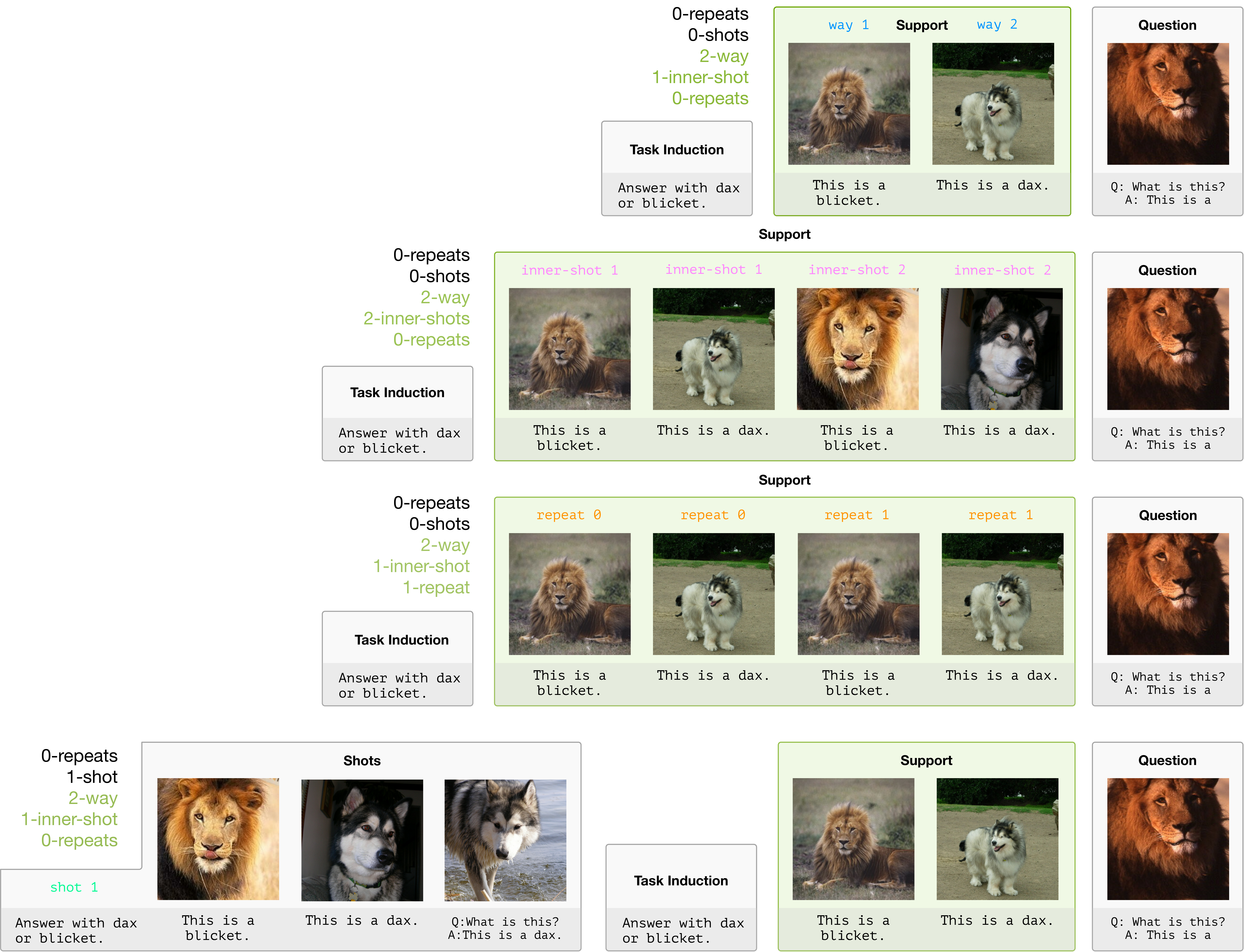}
\caption{Examples of few-shot learning vocabulary for fast-binding.}
\label{fig:vocab_extended_complete}
\end{figure}

As \Model can be conditioned on a sequence of interleaved images and text, it is capable not only of performing on a variety of multimodal tasks, but also, the same task can be induced in multiple ways to help \Model to learn and perform better. In order to make it easier to distinguish among these different ways of 'inducing' a task to the model, we have formalized the terminology used in our settings, which is described in section 3.4 of the main text. In \autoref{fig:vocab_complete} and \autoref{fig:vocab_extended_complete} below we provide more visual examples of this terminology.

\subsection{Tasks to Evaluate Fast-Binding Capacity} 

\subsubsection{Open-Ended MiniImageNet}

To construct the Open-Ended MiniImagenet evaluation we begin with the same subset $S$ of ImageNet classes applied in prior on meta-learning with MiniImagenet (See the appendix of~\cite{ravi2016optimization}). All images are taken from the validation set of ImageNet.

To generate a $2$-way question with $n$ inner-shots, the following process is followed: 

\begin{enumerate}
    \item Sample two classes $c_1, c_2$ from $S$
    \item Sample $n$ images $v^{c_1}_{1} \dots v^{c_1}_{n+1}$ from $c_1$ and $n$ images $v^{c_2}_{1} \dots v^{c_2}_{n}$ from $c_2$
    \item Interleave into a sequence of $2n$ support images $[v^{c_1}_1, v^{c_2}_1  \dots v^{c_1}_n, v^{c_2}_n] $
    \item Assign the nonsense words (\emph{dax}, \emph{blicket}) to $c_1, c_2$ at random, and interleave support captions \emph{"this is a dax"} or \emph{"this is a blicket"} accordingly
    \item Select one of $c_1, c_2$ at random, $c_q$, and sample a further question image  $v^{c_q}$ 
    \item Assign the truncated caption \emph{"this is a"} to $v_q$ and the appropriate nonsense word as the correct answer.
\end{enumerate}

Note that this process ensures that the image class and nonsense word assigned to the correct answer occur in either first or second place in the support, and the correct answer may be \emph{dax} or \emph{blicket} with equal probability.

To generate a $5$-way question, the above process is generalized. In 1. five distinct classes are sampled from $S$. The set of nonsense words applied in step 4. and 6 is: [\emph{dax, blicket, slation, perpo, shously}]. The final three words were taken from a nonsense-word generator\footnote{\url{https://www.soybomb.com/tricks/words/}} and selected because, like \emph{dax} and \emph{blicket} and for consistency, they decompose into two tokens in our model's subword vocabulary. 

All images are stored at $224 \times 224$ resolution. 

\subsubsection{Real-Name miniImageNet}

To generate Real-Name miniImagenet, the same process is followed, except that in steps 4. and 6., instead of using nonsense words to caption the support images (e.g. \emph{"this is a dax"}), the (first) class name from the ImageNet dataset is used (e.g.\ \emph{"this is a fruit bat"}).  

\subsubsection{Fast-VQA}

Unlike Open-Ended miniImageNet, Fast-VQA uses images from all 1,000 classes in the ImageNet dataset. For the evaluations in this paper, we again only take images from the validation set. Denote by $W$ the set of all 1,000 class (first) names, and for each $w_i \in W$, the corresponding set of images $c_i$. 

The Visual Genome (VG) dataset contains meta-data, questions and answers, such that we can consider data in the form $(Im, q, a, Ob)$, where $Im$ is the image, $q$ is the corresponding question, $a$ is the answer and $Ob$ is a list of names for all objects annotated in $Im$. We first filtered the dataset into a subset $VG*$ such that every question $q_k$ contained at least one word $w_i \in W$ \emph{and such that} the corresponding object list $Ob_k$ also contained $q_k$ and at least one other word $w_j \in W : w_j != w_i$. Thus, we can consider the elements of $VG*$ to be of the form $(Im, q, a, Ob, w_i, w_j)$

To generate a $2$-way, $n$-shot Fast-VQA question out of an element $(Im, q, a, Ob, w_i, w_j)$, we then did the following:

\begin{enumerate}
    \item Sample $n$ images $v^{c_i}_{1} \dots v^{c_i}_{n+1}$ from $c_1$ and $n$ images $v^{c_j}_{1} \dots v^{c_j}_{n}$ from $c_2$
    \item Depending on coin toss, form either the support $[v^{c_i}_1, v^{c_j}_1  \dots v^{c_i}_n, v^{c_j}_n] $ or the support $[v^{c_j}_1, v^{c_i}_1  \dots v^{c_j}_n, v^{c_i}_n]$
    \item Assign the nonsense words (\emph{dax}, \emph{blicket}) to $w_i, w_j$ at random, and interleave support captions \emph{"this is a dax"} or \emph{"this is a blicket"} accordingly
    \item Transform $q$ and $a$ into modified questions and answers $q*$ and $a*$ by replacing all instances of $w_i$ and any instances of $w_j$ with the corresponding strings \emph{dax} or \emph{blicket}
    \item Append the (VG) question $(Im, q*, a*)$ to the (ImageNet) support from 2. to create the Fast-VQA sample. 
\end{enumerate}

In this work, we only consider $2$-way Fast-VQA.

\subsubsection{Real-Fast-VQA}

To generate Real-Fast-VQA, the same process is followed, except that in step 3. the (first) class name from ImageNet is used to caption the support images (\emph{"this is a cat"}, \emph{"this is a wolf"}), and no string replacement is undertaken in 4.

\textbf{Links to download Open-Ended miniImageNet, Real-Name miniImageneNet, Fast-VQA and Real-Fast-VQA will be made available soon.}

\begin{figure}[ht!]
\includegraphics[width=\linewidth]{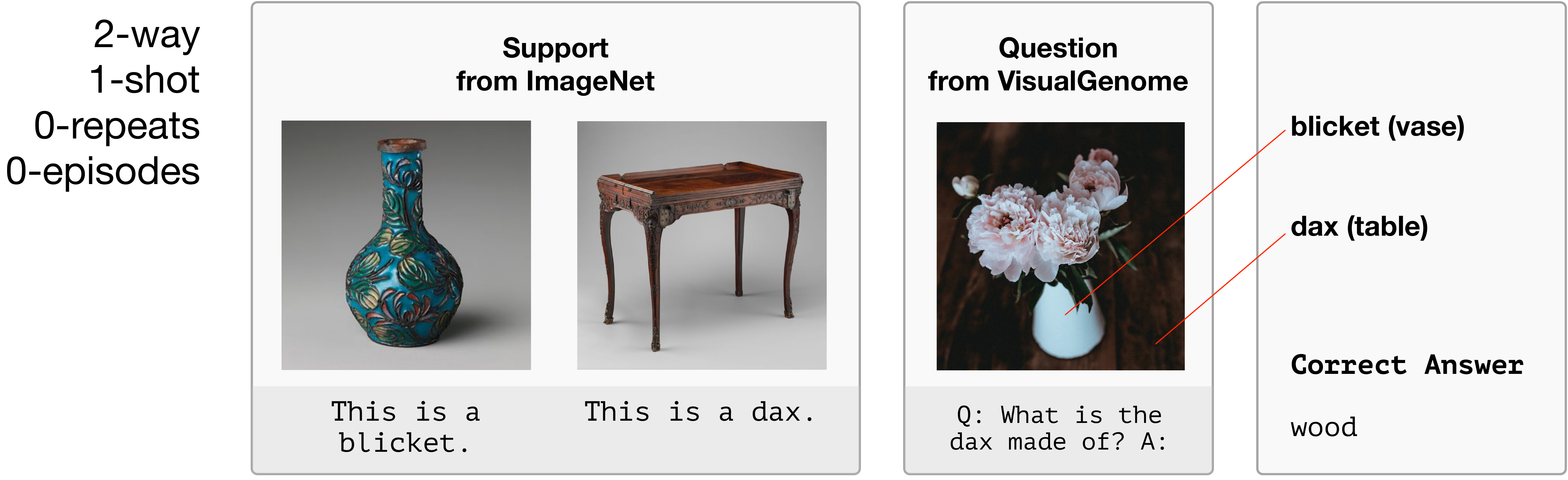}
\caption{Example of a Fast-VQA task.}
\label{fig:dax_vqa}
\end{figure}

\subsection{Encyclopedic Knowledge}

Here we add more detail to the claim in \autoref{sec:encyc_knowledge} that the model seems to be performing a sort of multi-hop deduction in the ``Wright Brothers'' example from \autoref{fig:headline}. 

\begin{figure}[ht!]
\centering
\includegraphics[width=\linewidth]{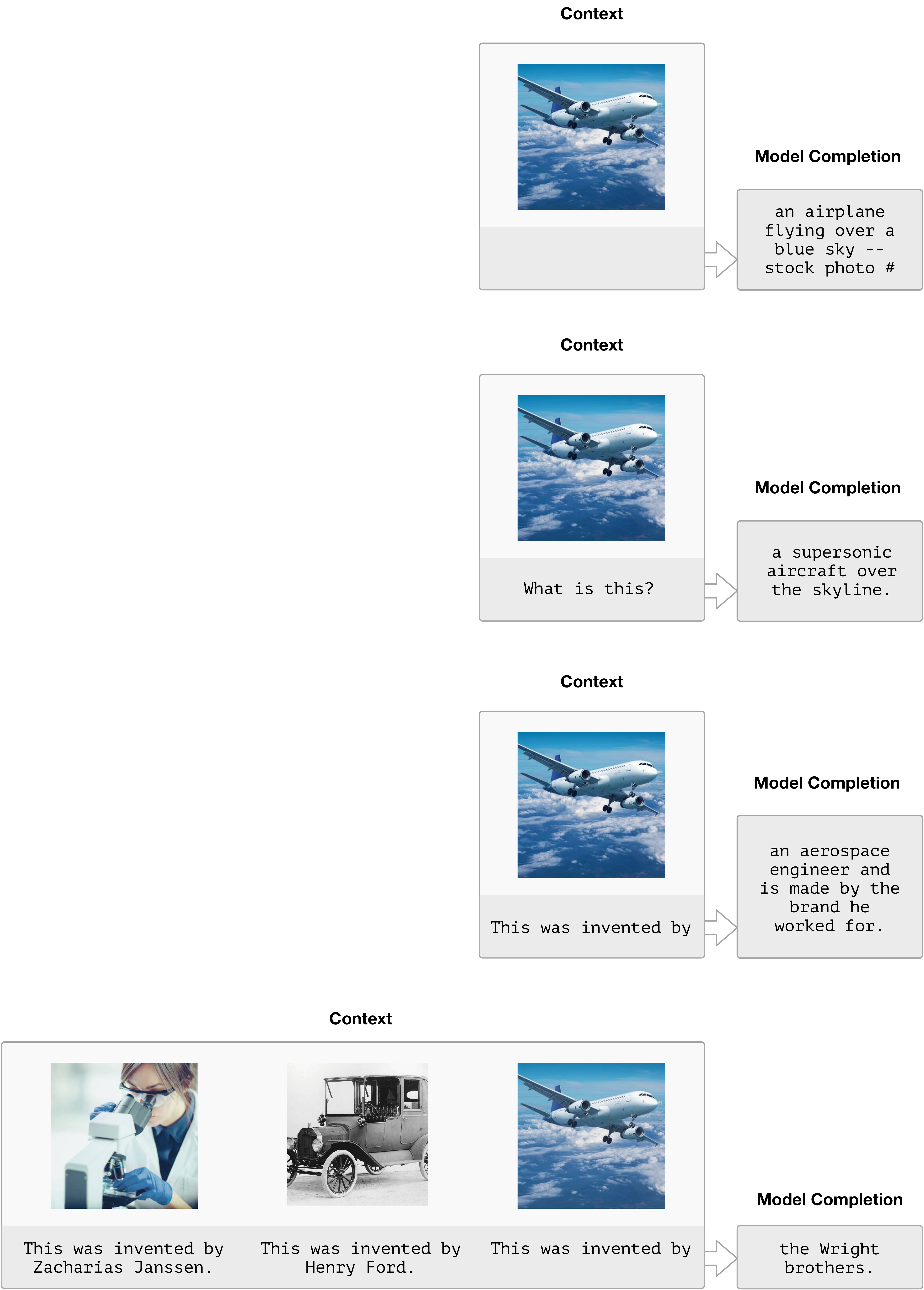}
\caption{Encyclopedic Knowledge. Shows the model retrieving factual knowledge given visual input. Required cherry-picking from around 5 seeds to get past common language model failure modes like simply repeating text from the prompt or emitting text that does not pertain to the test image.}
\label{fig:knowledge_wine}
\end{figure}

First, there has been a substantial amount of recent work studying a language model's ability to draw upon factual knowledge, examining the ability of language models to answer factual questions either zero-shot \cite{lama, brown2020language} or after open-domain QA finetuning \cite{roberts2020much, realmpaper, ragpaper}. Buoyed by these findings, we here demonstrate rigorously the impressive extent to which \Model seems to be commanding this factual knowledge and drawing upon it when prompted by an image (here an image of an airplane). We now break down why it is interesting that the model correctly determines that the Wright Brothers invented the object in the image (an airplane), by studying how the model responds to different prompts concerning this same test image in \autoref{fig:knowledge_wine}.

Recall that Conceptual Captions is \emph{hypernymed} so none of the language targets used to train \Model contain named entities like ``The Wright Brothers''. Instead, our training signal teaches the model to emit text that would roughly describe an image. The impressive finding is that this scalable, weakly supervised objective \emph{generalizes} to general information retrieval about an image. 

The top pane in \autoref{fig:knowledge_wine} shows an example of what the text in the captioning distribution looks like, captioning the image as ``an airplane flying over a blue sky -- stock photo \#''. Now, as established in \autoref{sec:rapid_adaptation} we enjoy some amount of zero-shot transfer from captioning to visual question-answering. This is demonstrated in the second and third rows of \autoref{fig:knowledge_wine}. But, adhering to the distribution of caption text, the model does not give a named entity when asked who invented the airplane. Instead it completes the prompt vaguely by saying ``This was invented by \emph{an aerospace engineer and is made by the brand he worked for}''.

But we know for certain that the language model has learned plenty of facts about named entities during pre-training and in particular we determined via the C4 dataset search tool \cite{c4_search} that there are multiple articles concerning the Wright Brothers. It's just that matching the distribution of Conceptual Captions text has taught the model to not emit named entities when prompted with an image. But the model can \emph{recover} the ability to refer to named entities given an image with few-shot learning (bottom row of \autoref{fig:knowledge_wine}). We show the model two examples of saying who invented an object depicted in an image by giving a named entity (Zacharias Janssen invented the microscope and Henry Ford invented the model T, an early automobile). With this prompt, \Model reliably retrieves the correct factual knowledge, having determined in the vision encoder that the image depicts an airplane, and having been demonstrated in-context that the desired output is the name of a person.

This outcome is robust, in the sense that we observed it in multiple versions of \Model during development, and in multiple examples, but drawing samples is not always successful and can require 3-4 tries to get past well-known language model failure modes of either repeating prompt text or emitting completely unrelated text. That's why we describe some samples as ``curated''.

We reiterate that this is a fascinating chain of deduction and a huge generalization leap from the task the model was trained to do, which is emit a caption for an image.

\begin{figure}[ht!]
\includegraphics[width=\linewidth]{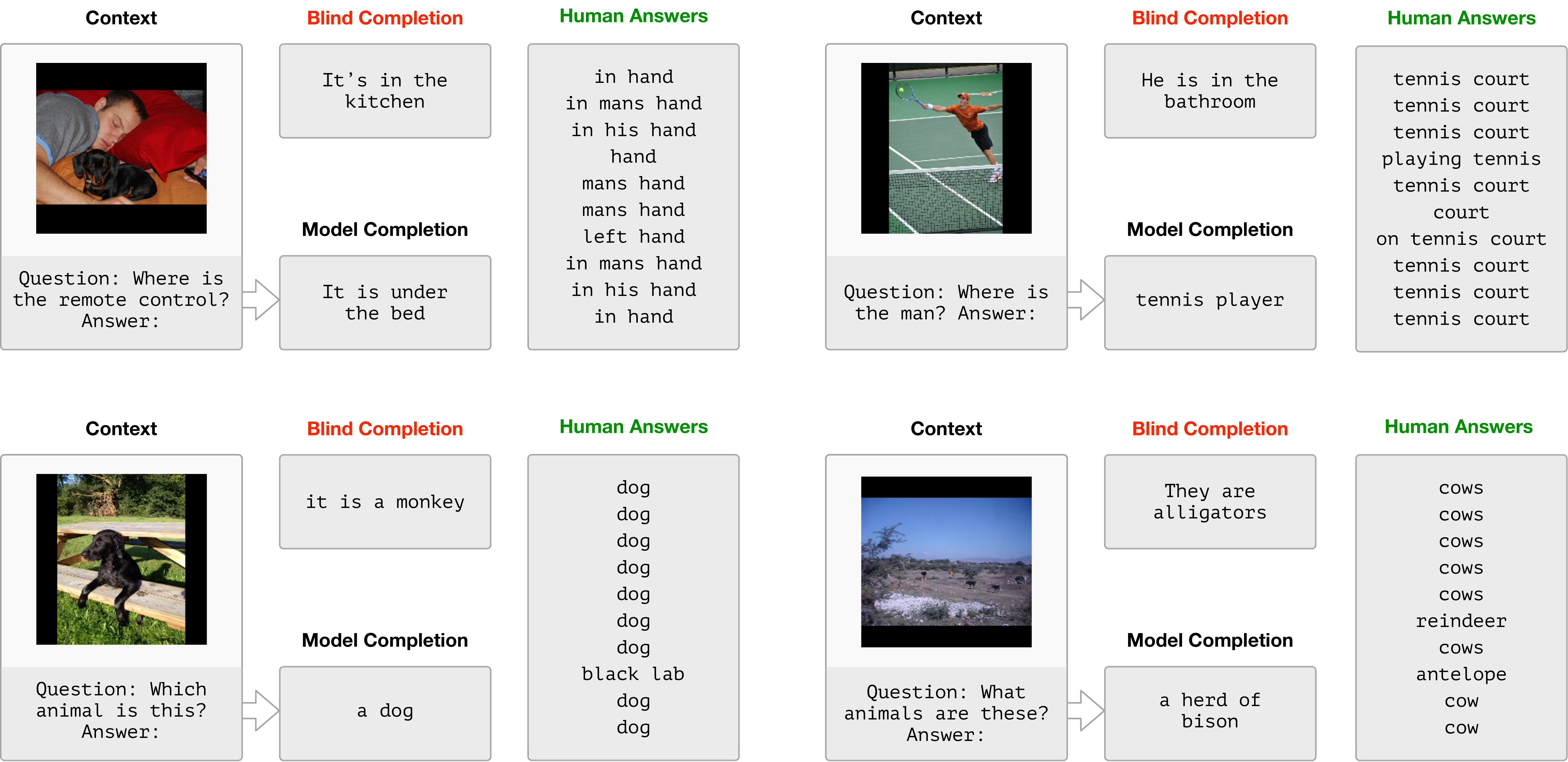}
\caption{VQA qualitative. This is a greedy sample of our model's prediction on a VQAv2 validation set example. See accuracy numbers in \autoref{tab:vqa_table} for overall robustness.}
\label{fig:vqa_qualitative}
\end{figure}

\begin{figure}[ht!]
\includegraphics[width=\linewidth]{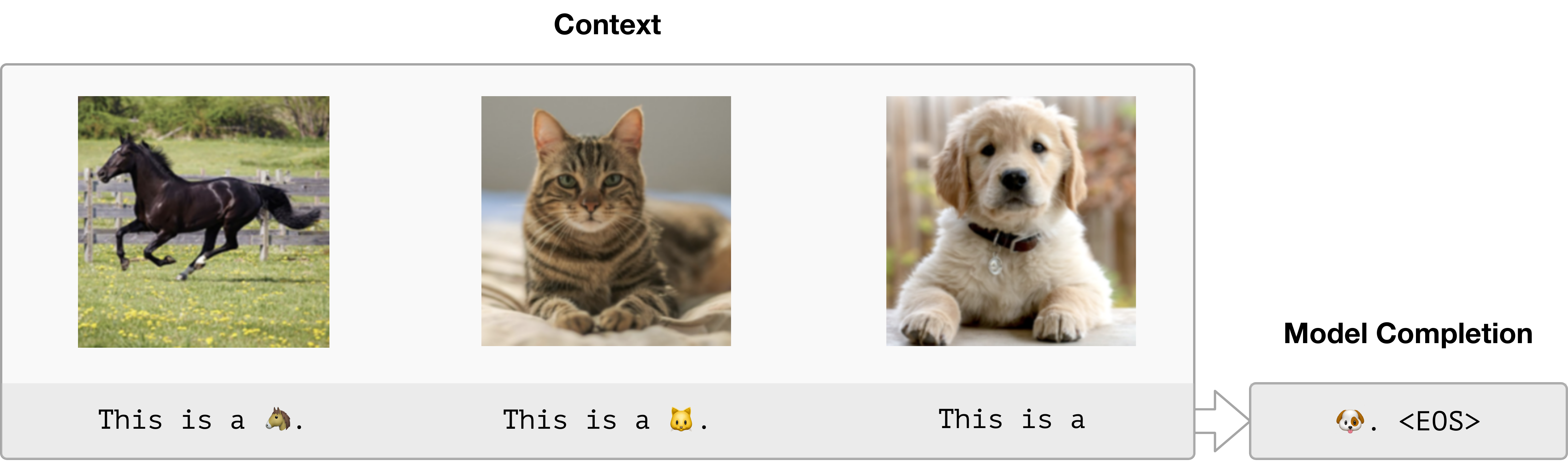}
\caption{Emojis. This sample reliably produced good output within a few attempts but did not work for every seed.}
\label{fig:knowledge_emoji}
\end{figure}

\begin{figure}[ht!]
\centering
\includegraphics[width=0.6\linewidth]{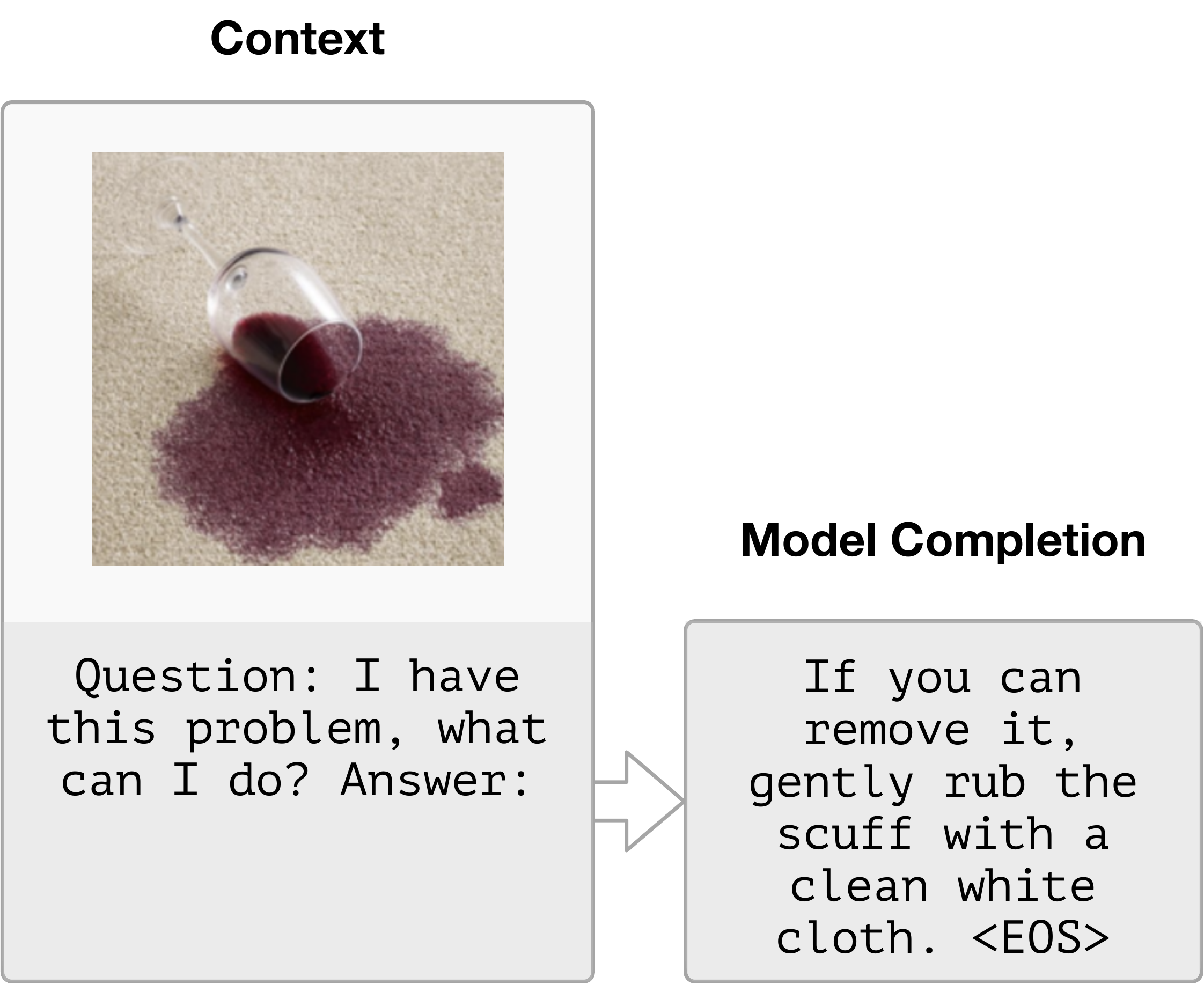}
\caption{Encyclopedic Knowledge. Demonstrates knowledge from language pre-training  being commanded given visual input. Required a few seeds to get a good answer which clearly paid attention to the image.}
\label{fig:wine_really}
\end{figure}

\end{document}